\documentclass[letterpaper, 10 pt, conference]{ieeeconf}  

\IEEEoverridecommandlockouts                              

\usepackage{multirow,booktabs}
\usepackage{graphics} 
\usepackage{epsfig} 
\usepackage{mathptmx} 
\usepackage{amsmath} 
\usepackage{amssymb}  
\usepackage{makecell}
\usepackage{bm}
\usepackage{colortbl}
\usepackage{hyperref}
\usepackage{xcolor}
\usepackage[most]{tcolorbox}
\makeatletter
\g@addto@macro\normalsize{%
  \setlength\abovedisplayskip{5pt plus 3pt minus 2pt}%
  \setlength\belowdisplayskip{5pt plus 3pt minus 2pt}%
  \setlength\abovedisplayshortskip{3pt plus 2pt minus 1pt}%
  \setlength\belowdisplayshortskip{3pt plus 2pt minus 1pt}%
}
\makeatother

\AtBeginDocument{%
  \setlength{\textfloatsep}{6pt plus 2pt minus 2pt}  
  \setlength{\floatsep}{6pt plus 2pt minus 2pt}      
  \setlength{\intextsep}{6pt plus 2pt minus 2pt}     

  \setlength{\dbltextfloatsep}{6pt plus 2pt minus 2pt}
  \setlength{\dblfloatsep}{6pt plus 2pt minus 2pt}

  \setlength{\abovecaptionskip}{3pt plus 1pt minus 1pt}
  \setlength{\belowcaptionskip}{0pt}
}

\title{\LARGE \bf
Perception-Aware Multimodal Spatial Reasoning from Monocular Images
}

\author{Yanchun Cheng, 
        Rundong Wang, 
        Xulei Yang, 
        Alok Prakash, 
        Daniela Rus, 
        Marcelo H Ang Jr, 
        ShiJie Li
\thanks{Y. Cheng, R. Wang and Marcelo H Ang Jr are with NUS, Singapore.
A. Prakash is with NTU, Singapore.
D. Rus is with MIT, US.
X. Yang, and S. Li are with the A*STAR, Singapore.
}%
}

\begin{document}

\maketitle
\thispagestyle{empty}
\pagestyle{empty}

\begin{abstract}
Spatial reasoning from monocular images is essential for autonomous driving, yet current Vision–Language Models (VLMs) still struggle with fine-grained geometric perception, particularly under large scale variation and ambiguous object appearance. We propose a simple yet effective perception-aware multimodal reasoning framework that equips VLMs with explicit object-centric grounding ability. Instead of relying on textual bounding-box outputs, each referred object is represented using all Visual Reference Tokens (VRTs) within its spatial extent, enabling visual evidence and textual reasoning to be processed jointly in a unified token space. To further strengthen cross-modal interaction, we construct a Multimodal Chain-of-Thought (MM-CoT) dataset that injects aligned visual and textual reasoning signals. A deterministic ordering strategy is introduced to make supervision over inherently unordered VRT sets fully compatible with the VLM’s autoregressive next-token prediction. With only standard supervised fine-tuning, our method achieves substantial improvements on the SURDS benchmark, outperforming previous approaches—including those using RL-based post-training—by a large margin across both single-object and multi-object tasks. These results demonstrate that accurate perception and multimodal reasoning are mutually reinforcing, and together form the key to robust spatial understanding in challenging monocular driving scenarios.
\end{abstract}

\section{INTRODUCTION}

Spatial reasoning, a fundamental capability for embodied AI, has attracted increasing research attention in recent years. It enables an agent to infer and understand 3D environments from purely 2D observations. Vision–Language Models (VLMs), equipped with strong perceptual and semantic reasoning abilities, therefore present promising potential for advancing this capability.

\begin{figure}[thpb]
   \centering
   \includegraphics[width=\linewidth]{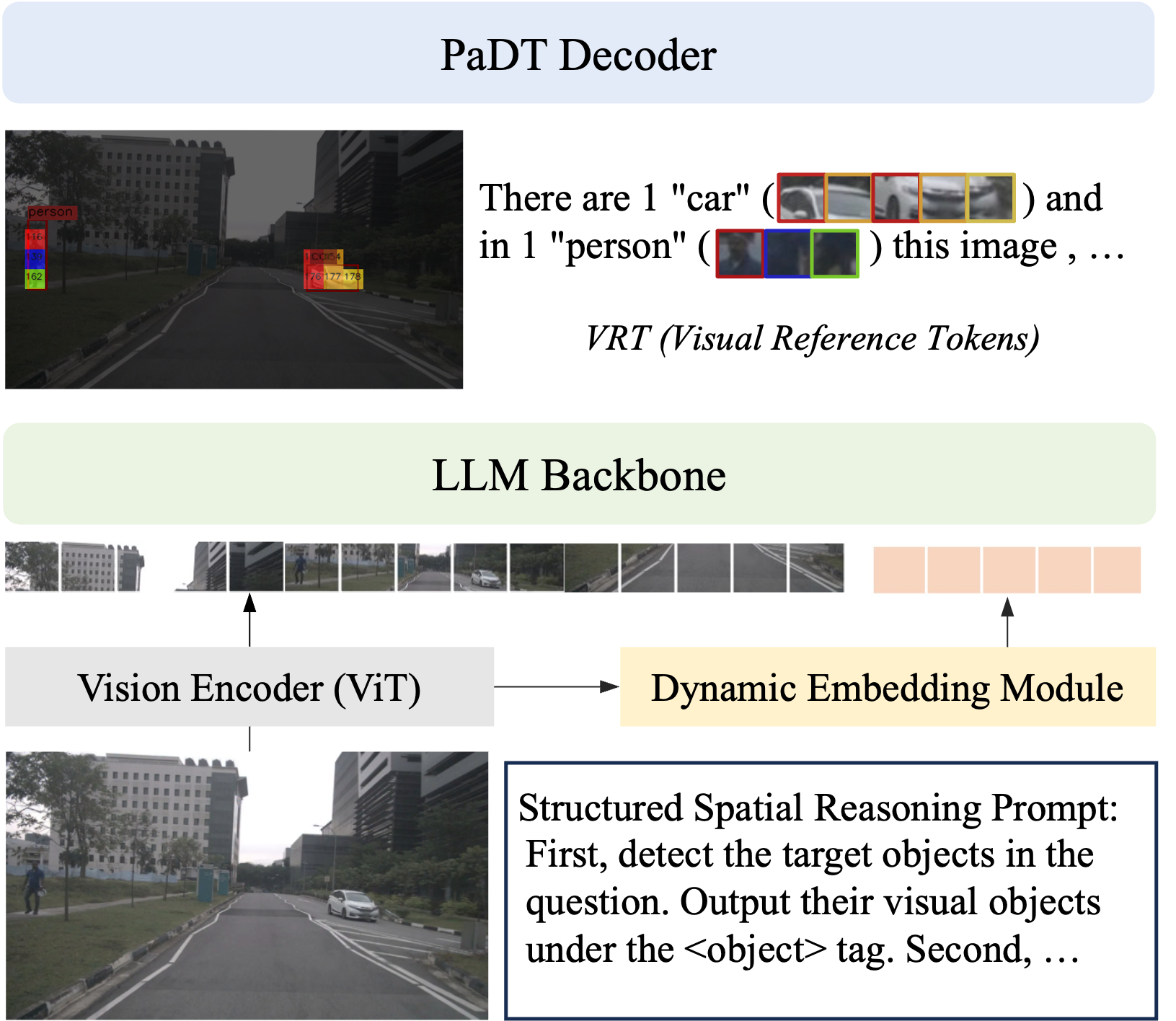}
   \caption{Overview of the perception-aware multimodal reasoning framework. Visual tokens from a ViT encoder are projected into the LLM, while a Dynamic Embedding Module injects object tokens and index tokens to enable explicit object-centric grounding. Grounding markers delimit visual reference spans, allowing the LLM to jointly reason over text, visual cues, and object instances. Right: examples of detection and grounding, region understanding, and grounded image conversation supported by the framework.}
   \label{framework}
\end{figure}

Recent spatial reasoning research primarily focuses on indoor environments and relies heavily on multi-view image inputs. However, such settings do not generalize well to outdoor scenarios, where autonomous driving serves as a representative and highly challenging application. In autonomous driving systems, a ring of cameras is typically deployed to provide multi-view coverage, yet the overlap between adjacent views is often minimal. Consequently, although multiple images exist, each view remains largely independent, making the problem effectively closer to a monocular setting. Moreover, due to monocular ambiguity and the substantially larger perception range in outdoor environments, object scales can vary dramatically in appearance, further complicating spatial reasoning. Some prior works attempt to address this issue by first locating the referred objects and then answering the question. Although this strategy yields certain performance improvements, these methods rely on text-based visual grounding, which lacks semantic expressiveness and often require costly reinforcement learning (RL)–based post-training procedures.

In this work, we follow a perception-then-answer paradigm and propose a simple yet efficient solution to improve VLMs spatial reasoning ability under a monocular setting. Specifically, each referred object is represented by all Visual Reference Tokens (VRTs)~\cite{su2025patch} whose centers fall within its spatial extent. Instead of predicting semantically meaningless textual bounding-box coordinates, our method is trained to predict these VRTs directly. Since VRTs reside in the same embedding space as textual tokens, both visual cues and textual reasoning naturally coexist and interact within a unified representation space. To further enhance this interaction and strengthen the model’s reasoning capability, we adopt the Chain-of-Thought (CoT) paradigm, which has demonstrated strong effectiveness across various applications. In particular, we construct a Multimodal CoT (MM-CoT) dataset in which textual and visual information are jointly encoded, and use it to fine-tune the proposed architecture.

In the proposed method, the referred objects are represented as an unordered subset of VRTs. However, VLMs inherently operate under a causal inference paradigm and generate outputs in a strictly ordered, sequential manner due to the next-token-prediction objective, even though their internal representations are order-agnostic. This fundamental mismatch prevents a direct application of autoregressive supervision to unordered VRT sets.
To resolve this issue, we follow the strategy adopted in Mamba-style sequence modeling and impose a deterministic ordering over all target-object VRTs. This enforced ordering establishes a one-to-one correspondence between the ground-truth VRT sequence and the predicted sequence, making the loss computation fully compatible with the VLM’s causal next-token prediction framework.

In our experiments, we observe that fine-tuning the proposed model on our self-constructed MM-CoT dataset using a simple supervised learning scheme already surpasses previous approaches by a large margin on the challenging monocular spatial reasoning task. This clearly demonstrates the effectiveness of our framework. Our contributions are summarized as follows:
\begin{itemize}
    \item We propose a perception-then-answer paradigm tailored for monocular spatial reasoning in challenging autonomous driving scenarios. In addition, our self-constructed MM-CoT dataset effectively encourages multimodal reasoning and significantly enhances the model’s overall reasoning capability.
    \item We present a solution that makes loss computation on unordered VRTs fully compatible with the VLM’s causal next-token prediction framework.
    \item Our method significantly outperforms prior approaches using only simple and efficient supervised fine-tuning, without relying on expensive RL-based post-training. This demonstrates both the effectiveness and the practicality of our design.
\end{itemize}

\section{Related Works}
In recent years, the remarkable advancements in large language models (LLMs)~\cite{r9,r57,r58} have stimulated significant research efforts directed toward extending natural language-based large models—particularly those within the GPT series of LLMs—into multimodal large language models (VLMs)~\cite{r1, r43, r44, r62, r70}. Within this domain, the integration of visual and linguistic modalities has witnessed substantial progress, leading to the development of numerous VLMs~\cite{r5, r6, r01, r02, r47, r48}. These models are applied to a variety of cross-modal applications such as visual question answering (VQA)~\cite{r3, r31}and cross-modal reasoning~\cite{r80, r28, r67}, facilitated by the accessibility of large-scale image-text datasets~\cite{r46, r39, r12}. Representative VLM architectures include the BLIP family~\cite{r43, r44}, the LLaVA family~\cite{r47, r48}, and the Qwen-VL family~\cite{r5,r6}. These models introduce innovations either in network architecture~\cite{r17,r60,r43,r44}or through novel training methodologies~\cite{r5,r83}. For instance, in terms of architectural design, Qwen-VL~\cite{r5} and MiniGPT-4~\cite{r83}utilize a Vision Transformer (ViT)~\cite{r2}-like network as the visual encoder, whereas LLaVA~\cite{r60} employs CLIP ViT-L/14~\cite{r63} and InternVL~\cite{r17} adopts InternViT-6B for visual encoding. Regarding training strategies, Qwen-VL~\cite{r5}implements a three-stage procedure: initial pre-training on large-scale image-text pairs, subsequent multi-task pre-training across seven core tasks, and final instruction tuning with over 350,000 dialogue instances. MiniGPT-4~\cite{r83}, on the other hand, follows a two-stage training scheme, beginning with pre-training on a composite dataset comprising Conceptual Captions~\cite{r13}, LAION~\cite{r66}, and SBU~\cite{r59}, followed by fine-tuning on a high-quality image description corpus.

In the era preceding the rise of LLMs, the majority of public vision-language datasets were oriented towards single tasks, which constrained their capacity to provide a holistic evaluation of multimodal reasoning capabilities. Notable examples of such benchmarks include image captioning~\cite{r46}, visual question answering~\cite{r3,r31}, and optical character recognition (OCR)~\cite{r50}. With the advent of LLMs, more comprehensive and multi-task benchmark datasets have been developed to better assess general-purpose multimodal reasoning. Among these, MME ~\cite{r26} emphasizes binary (yes/no) questions, visual perception, and linguistic reasoning; MMBench~\cite{r49} broadens the scope across diverse domains through a circular evaluation framework; Seed-Bench~\cite{r41,r42} incorporates multi-image and video inputs; and MMVet~\cite{r78} integrates multiple sub-tasks such as OCR, recognition, and mathematical reasoning. Beyond recognition-centric evaluations, recent initiatives aim to assess broader cognitive capabilities: MMMU~\cite{r79} focuses on reasoning involving domain-specific knowledge, HallusionBench~\cite{r33} investigates model hallucinations and visual illusions, MathVista~\cite{r52} targets mathematical reasoning in visual contexts, BLINK~\cite{r27} examines holistic perceptual understanding, and Mega-Bench~\cite{r14} scales evaluation to encompass over 500 real-world tasks.

While general-purpose MLLMs exhibit broad capabilities, their performance on fine-grained visual perception remains constrained~\cite{dehghani2023scaling, fang2023eva, wang2023internimage}. A key limitation arises from the dependence of vision encoders on fixed patch-based tokenization schemes, which can obscure local details and hinder performance in tasks requiring precise object localization, counting, or optical character recognition. To address this, adaptive image tiling methods—such as NaViT-style patch dropping and AnyRes~\cite{luo2023cheap, chen2024internvl, liu2024llava} improved—introduce flexibility by processing variable-resolution image tiles, thereby enhancing effective spatial resolution.

Concurrently, another research direction employs reinforcement learning to augment perceptual and reasoning abilities, as demonstrated by models including VLM-R1~\cite{shen2025vlmr1}, Visual-RFT~\cite{liu2025visual}, VisRL~\cite{chen2025visrl}, and Seg-R1~\cite{you2025seg}. These methods exhibit improved generalization and emergent skills such as segmentation and visual grounding. Although prior efforts have largely relied on reinforcement learning~\cite{chen2025visrl} or instruction tuning~\cite{jiang2024chatrex} to bolster visual reasoning, the use of learned queries as anchors for visual perception remains an underinvestigated area. Furthermore, the development of a unified architectural framework capable of seamlessly supporting diverse vision tasks continues to pose a significant open challenge.

A parallel research thrust aims to establish unified visual-linguistic representations through multi-granular tokenization strategies. At the regional level, various methods encode object bounding boxes or segmentation masks into geometric tokens \cite{chen2023shikra, xuan2024pink, peng2023kosmos, youferret} or learnable proxy embeddings \cite{zhang2024gpt4roi, yuan2024osprey, chen2023position, rasheed2024glamm}, typically anchored by detection frameworks or SAM \cite{kirillov2023segment}. This facilitates more accurate spatial grounding between visual and linguistic elements.

At the patch level, models like the Emu family \cite{sun2023emu} and LaVIT \cite{jin2024unified} leverage CLIP-derived patch embeddings as foundational visual vocabularies, enabling denser cross-modal alignment. Recent advancements further explore autoregressive quantization of image patches \cite{team2024chameleon, sun2024autoregressive}, transforming continuous pixels into discrete visual sequences that support efficient multimodal modeling. Finer-grained tokenization schemes are also emerging, as seen in \cite{ma2025clawmachine}.

While existing methods emulate linguistic structure through region-, instance-, or pixel-level discretization, they often fall short of achieving deep semantic integration between vision and language. To bridge this gap, we introduce a dynamic multimodal token space that establishes fine-grained correspondences between linguistic tokens and visual patches within a unified autoregressive modeling framework.

\section{Methodology}
\label{sec:method}

In this section, we provide a detailed description of the proposed method, which is designed for the challenging monocular spatial reasoning task in autonomous driving scenarios, where dramatic scale variations pose significant difficulties for existing approaches. Thus, the proposed method follows a perception-then-answer paradigm to strengthen its perceptual capability.
We begin with the necessary preliminaries in Sec. \ref{sec:pre}. An overview of the entire framework—including the task formulation and model architecture, is presented in Sec. \ref{sec:overview}. We then introduce the core component of our approach: the design and construction of the Multi-Modal Chain-of-Thought (MM-CoT) dataset. Finally, the learning objectives and supervision strategies are described in Sec. \ref{sec:sup}.

\subsection{Preliminary}
\label{sec:pre}

\textbf{Vision Language Model  (VLMs)}
 aim to jointly understand visual inputs and natural language by aligning image representations with textual embeddings in a unified semantic space. Modern VLMs typically consist of a visual encoder for extracting image features, a language model for processing textual instructions, and a multimodal fusion module that enables cross-modal interaction. Through large-scale pre-training on image–text pairs, these models acquire strong abilities in high-level scene understanding, instruction following, and open-ended reasoning. However, despite their impressive semantic understanding capabilities, most existing VLMs lack fine-grained geometric perception, making them inadequate for spatial reasoning tasks that require precise localization, depth interpretation, and object-level differentiation.

\textbf{Patch-as-decoable-token} 
Visual Reference Tokens (VRTs), introduced in PaDT \cite{su2025patch} , provide a unified patch-level visual representation for multimodal large language models. Instead of encoding object locations using textual bounding-box coordinates, VRTs are derived directly from the image’s patch embeddings and dynamically inserted into the model’s token vocabulary. Each VRT corresponds to a specific image region and carries both spatial and semantic information. By interleaving these visual tokens with textual tokens in the autoregressive sequence, VRTs establish an explicit and fine-grained link between language instructions and localized visual content.
This property makes VRTs particularly suitable for tasks that require precise object grounding or region-level reasoning. However, it remains an open question whether such a representation can effectively facilitate deeper understanding. In this work, we show that these visual tokens can indeed interact with textual information to enable multimodal reasoning, ultimately leading to significant improvements in the model’s reasoning capability.



\subsection{Overview}
\label{sec:overview}

\textbf{Task Formulation}
In this work, the proposed method is designed toward spatial reasoning under a monocular setting. 
Specifically, given a monocular image $I$ and a textual query $q$, the vision--language model $VLM$ 
is expected to output a correct answer $\hat{a}$, which can be formulated as:
\begin{equation}
    \hat{a} = VLM(I, q).
\end{equation}

The proposed method, shown in Fig.~\ref{framework}, follows a perception-then-answer paradigm that enhances the model’s understanding capability through improved perception, making it particularly suitable for the challenging monocular setting in autonomous driving scenarios. This paradigm assumes that the textual query typically refers to one or several key objects in the visual scene. Accordingly, the model first identifies and localizes these objects in the image, and then reasons over their spatial or semantic relationships to produce the final answer. Previous solutions require the model to directly output textual bounding-box representations, typically formatted as the coordinates of the four corner points. However, such a strategy suffers from weak semantic association between textual and visual information, as indicated by \cite{su2025patch}. In particular, bounding-box coordinates provide only geometric information and contain no semantic cues, making it impossible, for example, to infer the object’s category or distinguish between visually similar regions solely from their coordinate values.

Thus, in this work, we represent each referred object using the subset of visual referring tokens whose centers fall within the spatial extent of the object.
\begin{equation}
    Obj = VRT_N, \quad VRT_i \in Area(Obj)
\end{equation}
With such a representation, locating a referred object becomes equivalent to predicting a bunch of semantically meaningful token features, which can be directly generated by VLMs and naturally reside in the same semantic space as textual tokens. In this way, textual and visual information can be processed in a unified manner. Motivated by the effectiveness of the Chain-of-Thought paradigm, we construct a Multi-Modal Chain-of-Thought (MM-CoT) dataset to further encourage multimodal reasoning and strengthen the model’s overall reasoning capability.

\subsection{Multi-Modal Chain-of-Thought Format}
\label{sec:mm_cot}
Chain-of-Thought (CoT) data typically follows the format:

\begin{tcolorbox}[colback=gray!15, colframe=gray!40, boxrule=0.3pt]
\texttt{<think> ... </think>\\
<answer> ... </answer>}
\end{tcolorbox}

where the \texttt{<think>} segment contains intermediate textual reasoning steps, and the \texttt{<answer>} segment provides the final predicted answer. To construct such a dataset, the answer component typically adopts the ground-truth annotation, whereas the thinking component can originate from diverse sources. For example, the thinking component may be derived from model-generated Chain-of-Thought traces, human-written rationales, or existing reasoning datasets that provide step-by-step explanations. It is worth noting that our construction process is agnostic to the specific source of thinking annotations, and therefore can flexibly incorporate diverse forms of reasoning supervision.

To enable multimodal CoT, besides textual reasoning, we append a visual component represneted by VRTs that explicitly refers to specific image regions:
\begin{tcolorbox}[colback=gray!15, colframe=gray!40, boxrule=0.3pt]
\texttt{
<loc> VRT1, ... </loc>\\
<think> ... </think>\\
<answer> ... </answer>
}
\end{tcolorbox}

where \texttt{<loc>} encloses a set of Visual Referring Tokens (VRTs) corresponding to the referenced regions.

In our case, since we aim to let textual information interact with the visual evidence associated with the referred objects, the VRTs enclosed by \texttt{<loc>} are selected as those whose centers fall within the target object region. 

To further enhance the clarity of object references and support cases involving multiple referred objects, we additionally incorporate brief textual descriptions for each object. This is made possible by the unified processing of textual and visual tokens, which allows the model to jointly interpret these multimodal cues. The resulting multimodal CoT data sample is illustrated below:
\begin{tcolorbox}[colback=gray!15, colframe=gray!40, boxrule=0.3pt]
\texttt{
<loc> Obj1 refer to VRT1, ... \\
      ... \\
      ObjN refer to VRT1, ... </loc>\\
<think> ... </think>\\
<answer> ... </answer>
}
\end{tcolorbox}

\subsection{Supervision}
\label{sec:sup}
To enable both textual reasoning and visual grounding, our method employs two complementary
losses during training: a textual next-token-prediction loss and a PaDT-based visual grounding loss.

\textbf{Textual Loss.}
Following the standard next-token-prediction paradigm, we apply a cross-entropy loss over all
textual tokens. Given the autoregressive hidden state $h_t$ and the ground-truth textual token $y_t$,
the textual loss is defined as:
\begin{equation}
    L_{\text{text}} = - \log p(y_t \mid I, q, y_{<t}).
\end{equation}
This objective encourages the model to generate coherent reasoning steps and accurate answers
based on the input query and image.

\textbf{PaDT Loss.}
Different from the original PaDT formulation~\cite{su2025patch}, which randomly samples a
subset of foreground VRTs for supervision, our task requires using \emph{all} VRTs that belong
to the referred object. This provides complete visual grounding but also introduces a technical
challenge: although the set of object-related VRTs is inherently orderless, the VLM is trained in
an autoregressive (causal) manner and therefore expects the output tokens to follow a well-defined
sequence. Directly computing the loss over an unordered set leads to mismatched token alignment
and unstable optimization.

To address this issue, we follow the strategy used in Mamba-style sequence modeling and
impose a deterministic ordering over all target-object VRTs. This enforced ordering creates a
one-to-one correspondence between the ground-truth VRT sequence and the predicted sequence,
thus making the loss computation compatible with the VLM's causal next-token prediction.
Formally, let $S_{\text{obj}} = \{ v_1, \dots, v_K \}$ denote all VRTs whose centers
fall inside the target object. We sort these tokens using a predefined rule, producing an ordered sequence:
\begin{equation}
O_{\mathrm{obj}} = \mathrm{Order}(S_{\mathrm{obj}}).
\end{equation}

The VRT loss is then applied autoregressively:
\begin{equation}
L_{\mathrm{vrt}} = - \sum_{t=1}^{K} 
\log p\left( O_{\mathrm{obj}}[t] \mid I, q, y_{<t} \right).
\end{equation}
ensuring stable supervision and consistent alignment between textual and visual tokens.

\textbf{Overall Objective.}
The final training objective is a weighted sum of the two losses:
\begin{equation}
L = L_{\mathrm{text}} + L_{\mathrm{PaDT}}.
\end{equation}
This combined objective enables the model to learn textual reasoning, fine-grained visual
perception, and unified multimodal interaction within a single framework.

\begin{table*}[t]
\caption{Comparison results on visual spatial reasoning tasks.}
\label{tab:spatial_reasoning_results}
\centering
\scriptsize                      
\renewcommand{\arraystretch}{1.25}   
\resizebox{0.83\textwidth}{!}{
\begin{tabular}{lccccccccc}
\toprule
\textbf{Model} & \multicolumn{3}{c}{\textbf{Single-object}} & \multicolumn{3}{c}{\textbf{Multi-object}} & \textbf{Score} \\
\cmidrule(lr){2-4} \cmidrule(lr){5-7}
& Yaw & Pixel & Depth & Dis & L/R & F/B &  \\
\midrule
Random & 5.73 & 1.12 & 34.27 & 8.76 & 11.57 & 11.89 & 12.22 \\
GPT-4o & 13.08 & 1.62 & 2.49 & 11.57 & 47.89 & 3.14 & 13.30 \\
GPT-4o-mini & 3.24 & 0.28 & 0.22 & 4.22 & 21.51 & 2.05 & 5.25 \\
Gemini-1.5-pro & 19.14 & 4.41 & 22.70 & 61.95 & 66.38 & 22.05 & 32.77 \\
Gemini-2.0-flash & 9.30 & 5.41 & 32.97 & 69.30 & 77.30 & 20.00 & 35.71 \\
LLaVA-OV-Qwen2-72b-si & 1.95 & 3.03 & 23.57 & 3.78 & 9.73 & 8.65 & 8.45 \\
Qwen2.5-VL-72B-Instruct & 11.57 & 6.13 & 44.00 & 58.05 & 66.16 & 14.92 & 33.47 \\
Qwen2.5-VL-7B-Instruct & 7.57 & 3.46 & 25.95 & 11.46 & 17.95 & 9.30 & 12.61 \\
Qwen2.5-VL-3B-Instruct & 6.27 & 3.81 & 27.68 & 17.84 & 14.81 & 10.49 & 13.48 \\
\midrule
SpatialBot~\cite{cai2025spatialbot} & 0.00 & 0.00 & 12.00 & 0.00 & 0.00 & 0.00 & 2.00 \\
SpatialRGPT~\cite{SpatialRGPT} & 1.30 & 0.55 & 10.59 & 1.95 & 0.86 & 7.35 & 3.77 \\
\midrule
SURDS-3B & 20.97 & 44.81 & 69.84 & 49.30 & 51.35 & 8.54 & 40.80 \\
\midrule
Ours-3B & \textbf{49.11} & 19.23 & \textbf{95.39} & \textbf{77.59} & \textbf{87.46} & \textbf{79.64} & \textbf{68.07} \\
\bottomrule
\end{tabular}
}
\end{table*}

\begin{table*}[t]
\caption{Performance comparison of Qwen2.5-VL-3B variants on single-object and multi-object tasks.}
\label{tab:qwen25_vl_results}
\centering
\renewcommand{\arraystretch}{1.2}
\setlength{\tabcolsep}{8pt}
\begin{tabular}{c|cc|ccc|ccc|c}
\hline
\multirow{2}{*}{\textbf{Row}}& \multirow{2}{*}{\textbf{Init W}} & \multirow{2}{*}{\textbf{CoT}} &
\multicolumn{3}{c|}{\textbf{Single-object}} &
\multicolumn{3}{c|}{\textbf{Multi-object}} &
\multirow{2}{*}{\textbf{Score}} \\
\cline{4-9}
& & & \textbf{Yaw} & \textbf{Pixel} & \textbf{Depth}
& \textbf{Dis} & \textbf{L/R} & \textbf{F/B} & \\ 
\hline
1 & QWen & T & 6.27 & 3.81 & 27.68 & 17.84 & 14.81 & 10.49 & 13.48 \\
2 & QWen & MM & 23.54 & 1.27 & 24.88 & 54.69 & 68.24 & 29.42 & 33.67 \\
3 & PaDT & T & 27.95 & 18.31 & 93.19 & 53.19 & 79.14 & 54.32 & 54.35 \\
4 & PaDT & MM & 49.11 & 19.23 & 95.39 & 77.59 & 87.46 & 79.64 & 68.07 \\
\hline
\end{tabular}
\end{table*}

\section{Experiments}
\label{sec:experiments}

\subsection{Experimental Setting}

\noindent \textbf{Dataset} SURDS\cite{surds} is a large-scale spatial understanding dataset built upon the nuScenes driving corpus. After applying a multi-stage filtering pipeline—including occlusion removal, edge and size constraints, and description-based ambiguity filtering—the benchmark retains 27,152 training and 5,919 validation images, from which 41,080 training and 9,250 evaluation VQA pairs are constructed. SURDS covers six spatial reasoning tasks, including yaw orientation, pixel-level localization, depth range estimation, pairwise distance, left–right ordering, and front–behind relations, offering the first fine-grained spatial reasoning benchmark tailored for realistic driving scenarios.

\noindent \textbf{Implementation Details}
For a fair comparison, when constructing the MM-CoT dataset, the textual reasoning traces are directly adopted from SURDS \cite{surds}, since it serves as our primary baseline. In training, we initialize our model using PaDT \cite{su2025patch}, which shares the same architecture as Qwen2.5-VL but is further fine-tuned on an open-vocabulary object detection task. This initialization equips our method with strong fine-grained visual perception. We train the model with a learning rate of $1\times10^{-5}$ under a constant scheduler, using 8 NVIDIA H100 GPUs, and the full training process takes approximately 47 GPU-hours.

\noindent \textbf{Evaluation Metrics}
We follow SURDS benchmark experimental setting. Specifically, For the Pixel Localization Estimation task, a centerness-based metric following~\cite{wang2024cogvlm} is used. For all other tasks, each prediction is assigned a score of 1 if it matches the ground-truth answer and 0 otherwise. Given $N$ QA pairs, the score for each task is computed as the average accuracy over all $N$ samples. The overall score is then obtained by averaging the individual task scores.

\begin{figure*}[t]
    \centering
    \includegraphics[width=0.95\linewidth]{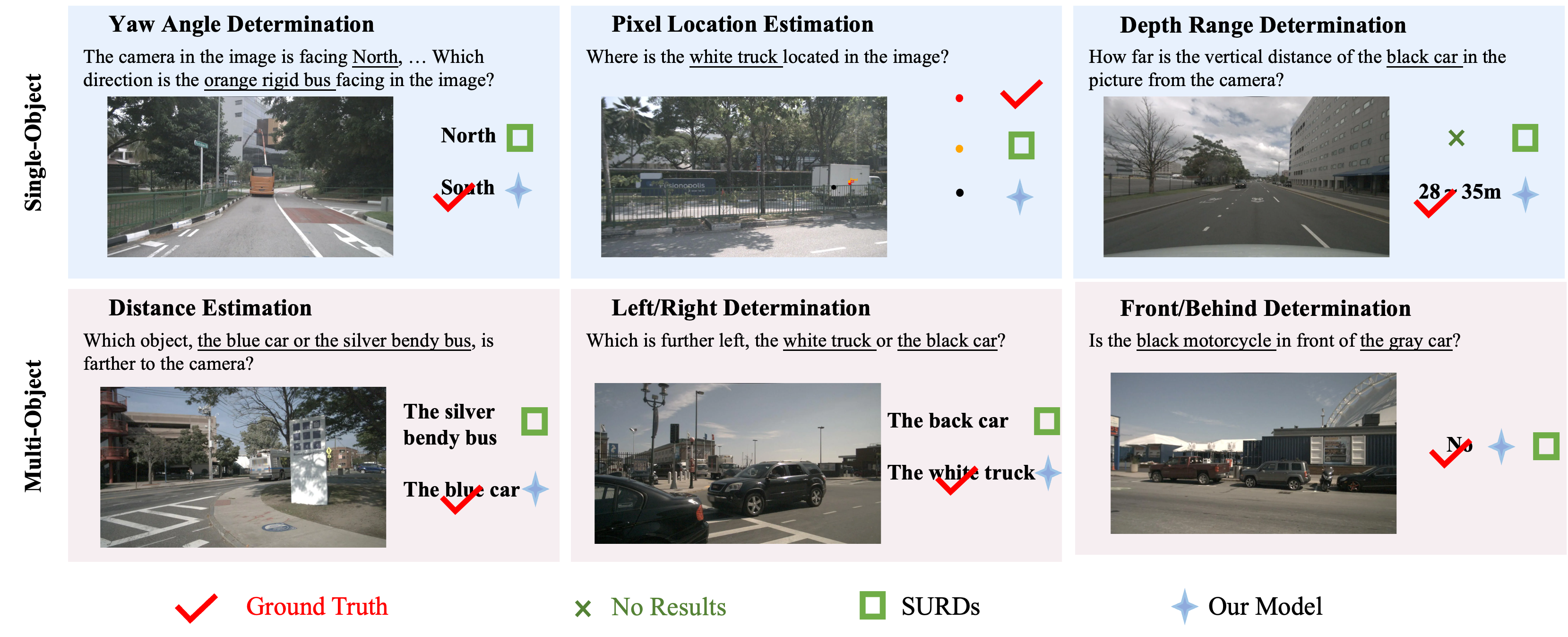}
    \caption{Illustrative examples of the benchmark QA pairs on both single-object and multiobject}
    \label{fig:viz}
\end{figure*}

\subsection{Comparison to the state-of-the-art}
The results in Tab. \ref{tab:spatial_reasoning_results} clearly demonstrate the effectiveness of our approach across all six spatial reasoning tasks. Compared with both general-purpose VLMs and spatially specialized baselines, our model achieves the highest overall score (68.07), outperforming the second-best system by a large margin. The improvements are especially pronounced in the single-object tasks, where our method achieves 49.11 on yaw angle determination and an exceptional 95.39 on depth estimation—far exceeding all competing models. These results indicate that our model is able to robustly capture absolute geometric properties that most VLMs struggle with, including large proprietary models such as GPT-4o and Gemini-2.0-flash. Although our model relies solely on standard supervised fine-tuning (SFT), it still outperforms methods that employ RL-based post-training strategies, further demonstrating its effectiveness.

A key reason behind these gains is our perception-first strategy: before answering any question, the model explicitly predicts the image patch corresponding to the target object. This step forces the model to ground its reasoning on localized visual evidence rather than relying on global heuristics or language priors. As a result, the subsequent reasoning stage operates on a correctly identified object region, enabling far more stable estimation of orientation, depth, and positional attributes—tasks where general VLMs typically fail due to inaccurate or missing object grounding.

The performance gains extend to multi-object relational reasoning, where our method achieves 77.59 on distance comparison and 87.46 on left–right ordering, again setting new best results. Even on the more challenging front–behind task, where many models collapse toward near-random behavior, our method retains competitive performance. By grounding each referenced object through predicted patches, the model is able to construct explicit spatial relationships between localized regions rather than implicitly inferring them from the entire image. These consistent improvements across both absolute and relational subtasks highlight the strength of our approach in handling diverse forms of spatial reasoning, and further confirm that accurate target localization is a crucial prerequisite for reliable spatial understanding. This also demonstrates that, in multi-object scenarios, VRT-based representations can distinguish between different objects much more effectively than purely textual descriptions.

\textbf{Qualitative Results}. We present qualitative visualizations to further illustrate the effectiveness of our approach, as shown in {Fig. \ref{fig:viz}. These examples demonstrate that the model can accurately locate referred objects and perform reliable spatial reasoning across diverse scenarios.

\subsection{Ablation Study}

Tab.~\ref{tab:qwen25_vl_results} presents the performance under different initialization settings. QWen corresponds to QWen-2.5-VL-3B, whereas PaDT~\cite{su2025patch} shares the same architecture but is further fine-tuned with an open-vocabulary detection objective, providing significantly stronger object-centric perception. This distinction allows us to isolate the role of perception priors: PaDT delivers substantial gains across all tasks even with purely textual CoT supervision (e.g., +21.68 Yaw, +14.50 Pixel, +65.51 Depth over QWen-T), indicating that robust perception is the foundation of effective spatial reasoning in monocular settings. Beyond perception, multimodal CoT (MM-CoT) further enhances performance for both QWen and PaDT. Unlike textual CoT, which only guides linguistic reasoning, MM-CoT enables the model’s thinking process to directly interact with visual reference tokens, allowing textual and visual information to influence each other within a unified representation space. This multimodal interaction leads to richer geometric understanding and yields the best results with PaDT-MM (e.g., 49.11 Yaw, 95.39 Depth, 79.64 F/B, 68.07 overall score). Together, these results demonstrate that high-quality perception and multimodal reasoning are complementary: PaDT provides strong perceptual priors, while MM-CoT enables effective cross-modal reasoning, jointly producing substantial improvements across both single-object and multi-object spatial reasoning tasks.

 \section{Future Work \& Limitation}
Although our method achieves strong performance with simple supervised fine-tuning, it does not leverage RL-based post-training, which has shown potential for improving long-horizon reasoning and exploration. A promising direction for future work is to investigate how RL-based optimization can further enhance multimodal reasoning and better utilize visual feedback, potentially leading to more robust spatial understanding in complex driving scenarios.

 \section{Conclusion}
In this work, we presented a perception-then-answer framework that significantly enhances monocular spatial reasoning in autonomous driving scenarios. By representing each referred object with its associated VRTs, our method replaces semantically weak text-based grounding with a unified visual–textual representation that enables richer multimodal interaction. The proposed MM-CoT dataset further strengthens this interaction by allowing the reasoning process to operate directly over both modalities. To address the mismatch between unordered VRT sets and the causal nature of autoregressive VLMs, we introduced a deterministic ordering mechanism that enables stable and fully compatible supervision.
Extensive experiments on the SURDS benchmark demonstrate that both components—strong perception priors (PaDT) and multimodal reasoning (MM-CoT)—are essential and complementary. Notably, our approach surpasses prior methods by a large margin using only simple supervised fine-tuning, without relying on costly RL-based post-training. This highlights the importance of accurate perception and cross-modal reasoning in advancing spatial understanding for real-world autonomous driving.



\bibliographystyle{IEEEtran}
\bibliography{IEEEabrv,refer}

\end{document}